\documentclass[letterpaper, 10 pt, conference]{ieeeconf}  

\IEEEoverridecommandlockouts                              

\usepackage{graphics} 
\usepackage{epsfig} 
\usepackage{amsmath} 
\usepackage{graphicx}
\usepackage{amsfonts}
\usepackage{mathrsfs}
\usepackage[dvipsnames,svgnames]{xcolor}
\usepackage{multicol}
\usepackage[font=footnotesize,labelfont=bf]{caption}
\usepackage{subcaption}
\usepackage{siunitx}
\usepackage{xspace}
\usepackage{amsbsy} 
\usepackage{dsfont}
\usepackage{booktabs}

\usepackage{geometry}
\usepackage{algorithm}
\usepackage[noend]{algpseudocode}

\let\labelindent\relax
\usepackage{enumitem}

\usepackage[symbol]{footmisc} 

\usepackage[dvipsnames]{xcolor}
\usepackage{hyperref}
\hypersetup{
    colorlinks=true,          
    linkcolor=red,            
    citecolor=blue,           
    urlcolor=green            
}
\hypersetup{
    linkbordercolor=blue,     
    citebordercolor=magenta,  
    urlbordercolor=cyan       
}

\newcommand{\be}{\begin{enumerate}}
\newcommand{\ee}{\end{enumerate}}

\newcommand{\bi}{\begin{itemize}}
\newcommand{\ei}{\end{itemize}}

\newcommand{\tem}{\textemdash}

\usepackage[T1]{fontenc}
\newcommand{\algo}[1]{\textsc{#1}}

\newcommand{\Mmap}{\ensuremath{\mathcal{M}}\xspace}
\newcommand{\Pmap}{\ensuremath{\mathcal{M}^{\mathrm{C}}}\xspace}

\newcommand{\Nmap}{\ensuremath{\mathcal{M}^{\mathrm{NC}}}\xspace}

\newcommand{\xyplane}{$xy$\nobreakdash-plane}

\newcommand{\Erobot}{\ensuremath{\mathbb{E_\mathsf{R}}}\xspace}
\newcommand{\Esensor}{\ensuremath{\mathbb{E_\mathsf{S}}}\xspace}
\newcommand{\Ejoint}{\ensuremath{\mathbb{E_\mathsf{J}}}\xspace}

\newcommand{\Crobot}{\ensuremath{c^\mathsf{R}}\xspace}
\newcommand{\Trobot}{\ensuremath{\pi_\mathsf{R}}\xspace}
\newcommand{\TranR}{\ensuremath{\mathbb{T}_\mathsf{R}}\xspace}
\newcommand{\Grobot}{\ensuremath{\mathbb{G}_\mathsf{R}}\xspace}

\newcommand{\Csensor}{\ensuremath{c^\mathsf{S}}\xspace}
\newcommand{\Tsensor}{\ensuremath{\pi_\mathsf{S}}\xspace}
\newcommand{\TranS}{\ensuremath{\mathbb{T}_\mathsf{S}}\xspace}
\newcommand{\Gsensor}{\ensuremath{\mathbb{G}_\mathsf{S}}\xspace}

\newcommand{\Cjoint}{\ensuremath{c^\mathsf{J}}\xspace}

\newcommand{\TranJ}{\ensuremath{\mathbb{T}_\mathsf{J}}\xspace}
\newcommand{\Gjoint}{\ensuremath{\mathbb{G}_\mathsf{J}}\xspace}

\newcommand{\Hsensor}{\ensuremath{H^{\psi}}\xspace}

\newcommand{\sStart}{\ensuremath{s_\mathit{start}}\xspace}
\newcommand{\sGoal}{\ensuremath{s_\mathit{goal}}\xspace}

\newcommand{\Splash}{\textsc{SPlaSH}}
\newcommand{\Split}{\textsc{SPLIT}}
\newcommand{\SplashVerb}{\textbf{S}ensor \textbf{Pla}nning with \textbf{S}ensor \textbf{H}istory (\Splash{})}
\newcommand{\SplitVerb}{\textbf{S}ensor \textbf{P}lanning with \textbf{L}ocal \textbf{I}terative \textbf{T}unneling (\Split{})}

\title{
    \LARGE
    \bf
        Search-based Planning for Active Sensing \\
        in Goal-Directed Coverage Tasks
}
\author{}
\author{
    Tushar Kusnur, Dhruv Mauria Saxena, and Maxim Likhachev
    \thanks{All authors are with the Robotics Institute, Carnegie Mellon University, Pittsburgh, USA
            {\tt\small \{tkusnur, dsaxena, maxim\}@cs.cmu.edu}.}%
}

\begin{document}
\setlength{\abovedisplayskip}{4pt}
\setlength{\belowdisplayskip}{4pt}
\maketitle
\thispagestyle{empty}
\pagestyle{empty}
\begin{abstract}
Path planning for robotic coverage is the task of determining a collision-free robot trajectory that observes all points of interest in an environment.
Robots employed for such tasks are often capable of exercising active control over onboard observational sensors during navigation.
In this paper, we tackle the problem of planning robot and sensor trajectories that maximize information gain in such tasks where the robot needs to cover points of interest with its sensor footprint.
Search-based planners in general guarantee completeness and provable bounds on suboptimality with respect to an underlying graph discretization.
However, searching for kinodynamically feasible paths in the joint space of robot and sensor state variables with standard search is computationally expensive.
We propose two alternative search-based approaches to this problem.
The first solves for robot and sensor trajectories independently in decoupled state spaces while maintaining a history of sensor headings during the search.
The second is a two-step approach that first quickly computes a solution in decoupled state spaces and then refines it by searching its local neighborhood in the joint space for a better solution.
We evaluate our approaches in simulation with a kinodynamically constrained unmanned aerial vehicle performing coverage over a 2D environment and show their benefits.
\end{abstract}
\section{Introduction}\label{sec:introduction}
Path planning for traditional robotic coverage is the task of determining a collision-free robot trajectory that observes all points of interest in a given environment~\cite{galceran2013survey}.
Numerous real-world tasks including environmental exploration, traffic monitoring, and post-disaster assessment can be cast as coverage problems~\cite{nedjati2016complete,smith2011persistent,srinivasan2004airborne,teixeira2018autonomous}.
Robots employed for such coverage tasks are often equipped with limited-range sensors to observe the environment and can exercise active control over them.
An important problem is to plan robot and sensor trajectories that maximize coverage or information gain in these tasks.
Planning in real time for kinodynamically constrained robots is in itself a computationally expensive problem due to the many degrees of freedom in a kinodynamic state.
Additionally planning trajectories for sensors onboard these robots further increases the computational complexity of this task.

In this paper, we consider the specific problem of planning trajectories for an unmanned aerial vehicle (UAV) and its onboard sensor covering cells of a discrete map\tem representing a known, deterministic environment\tem to achieve efficient 2D area coverage while navigating to an assigned goal.
We assume that the UAV flies at a fixed altitude, and that the yaw angle of a pan-only camera onboard the UAV can be controlled, thereby controlling the camera's projected footprint on the ground.
In turn, we include the robot's $x$ and $y$ coordinates on the map, heading angle $\theta$, velocity $v$, timestamp $t$, and sensor angle $\psi$ in our state space\tem making it \emph{at least} a 6-degree-of-freedom (6-DoF) planning problem (we describe in Sec.~\ref{sec:problem-and-notation} how this can amount to planning in more than 6 DoFs).

We tackle this problem using a search-based approach.
To the best of our knowledge, no previous work applies search-based planning to compute trajectories for a kinodynamically constrained robot and an onboard directional sensor to maximize coverage.
We propose two approaches to do so, each with a different way to improve the solution.
\begin{figure}[t]
    \centering
    \includegraphics[width=\columnwidth]{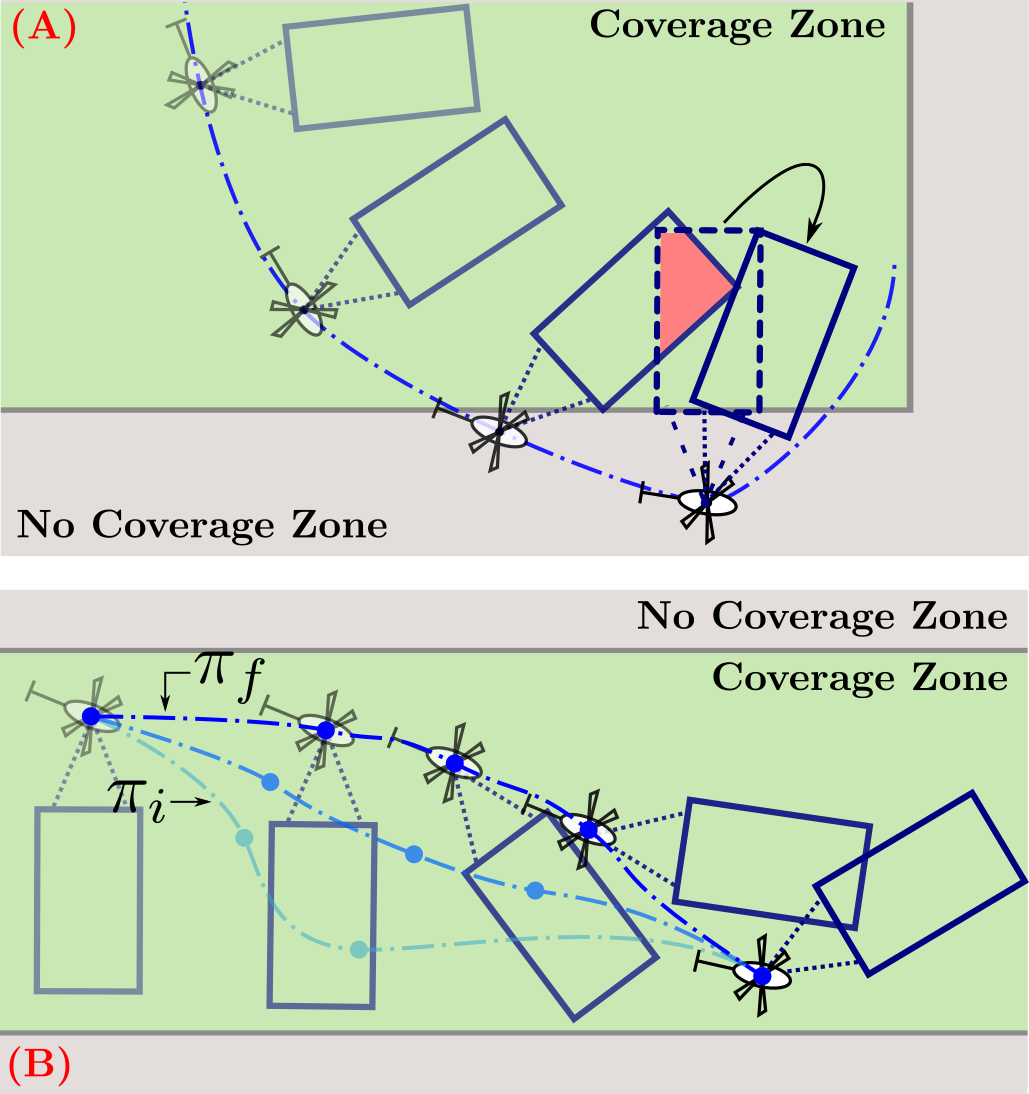}
    \caption{We present two algorithms for search-based planning for active sensing. (A) \Splash{} tries to minimize sensor footprint overlaps along the robot trajectory. For the last state in the figure, it would prefer the solid blue sensor footprint over the dashed blue so as to avoid the overlap denoted by the red area. (B) \Split{} iteratively refines an initial trajectory $\pi_i$ to maximize the area covered by the sensor to come up with the final trajectory $\pi_f$. }
    \label{fig:toy-example}
\end{figure}

A key challenge in planning non-myopic sensor trajectories that maximize coverage is that in general, for a given robot trajectory, the optimal sensor configuration at a given point in the trajectory depends on all previous sensor measurements (the full \emph{sensor history}) before it.
One can appreciate this in 2D environments by thinking of sensor footprint overlaps\tem to compute an optimal sensor configuration at a given location, a planner must take into account all overlaps with previously planned sensor footprints in the plan being considered.
However, including the full sensor history in a state makes the search computationally intractable.
Our first approach, \SplashVerb{}, first computes a robot trajectory.
It then searches for sensor plans for this fixed robot trajectory while maintaining a \emph{partial} sensor history during search\tem this prevents covering areas already sensed before in the trajectory, up to the size of the maintined sensor history.
Even for a fixed robot trajectory, the space of sensor trajectories exponentially increases in dimensionality as we account for larger sensor histories (this might not be trivial to see, and so we detail how this occurs in Sec.~\ref{sec:approach-one}).
Our results show that in most planning problems considering 2D sensor footprint overlaps, only a partial history of sensor footprints actually affect computing the next optimal sensor configuration.

The basis of our second approach is the empirical observation that approximate search algorithms like Weighted A* \algo{(WA*)}~\cite{pohl1970heuristic} overlook better solutions that are actually ``close'' to the computed solution in the space of solution paths~\cite{furcy2006itsa}.
In our second approach, \SplitVerb{}, we \emph{split} the process into two steps:
\begin{itemize}[noitemsep,topsep=0pt,leftmargin=13pt]
    \item
    We first quickly compute suboptimal robot and sensor trajectories in \emph{decoupled} robot- and sensor-state spaces (\textbf{initialization}).
    \item
    We then use this solution as initialization to a local-search routine that iteratively improves this solution in the \emph{joint} robot- and sensor-state space until time runs out (\textbf{refinement}).
\end{itemize}
We adapt Iterative Tunneling Search with \algo{A*} (\algo{ITSA*})~\cite{furcy2006itsa} to our problem for the refinement step.
This is detailed in Sec.~\ref{sec:approach-two}.

The illustation in Fig.~\ref{fig:toy-example} depicts both of our approaches on a toy example.
Our approaches can be contextualized within several related works on sensor management and informative path planning, as we do in Sec.~\ref{sec:related-work}.
In Sec.~\ref{sec:problem-and-notation}, we describe our problem and notation in detail.
In sections \ref{sec:approach-one} and \ref{sec:approach-two}, we describe and provide pseudocode for both of our approaches.
In Sec.~\ref{sec:results}, we evaluate our approaches and show their benefits in the context of a previously established planning framework for persistent coverage with multiple UAVs~\cite{kusnur2019planning}, where individual UAVs are tasked with generating collision-free trajectories that maximize information gain while navigating to an assigned goal.
Note that our contribution lies in planning robot and sensor trajectories for a single UAV navigating to a goal (we do not attempt to solve the problem of coordinating UAV plans for coverage).
\section{Related Work}\label{sec:related-work}
Hero and Cochran present an extensive survey on sensor management~\cite{hero2011sensor}.
Gutpta et al. state general challenges and computational complexity of optimal sensor selection in detail in~\cite{gupta2006stochastic}\tem this is on similar lines as the computational challenge of maintaining a sensor history (see Fig~\ref{fig:history}, explained later).
In general, optimal coverage has been addressed in various settings including mobile sensors and autonomous robots.
Robotic sensing systems have used with both fixed sensors~\cite{huang2017visual,furgale2010visual} as well as sensors that execute pre-computed patterns~\cite{scherer2012autonomous}.
The problem of optimal mobile sensor location with unbounded ranges has been tackled as Voronoi space partitioning in~\cite{du1999centroidal}.
Many approaches have also been targeted to specific applications, such as active perception work ~\cite{mori1990active,costante2016perception}.
The measurement control problem, also essentially a sensor scheduling problem, was shown to be solved by tree-search in general~\cite{meier1967optimal}.
To deal with computational intractability, several greedy solutions have been proposed~\cite{gupta2006stochastic,oshman1994optimal,mukai1996active,chung2004decentralized,kagami2006sensor}.
Further, Finite-horizon model predictive control provide improvement over myopic techniques but suffer from high run-times in large state spaces and provide no performance guarantees beyond the horizon depth~\cite{bourgault2003coordinated,ryan2010particle}.
Arora et al. propose a data-driven approaches to sensor trajectory generation that map calculated features to sensory actions~\cite{arora2015pasp}.

Several recent works that fall under \emph{informative path planning} are closely related to our work.
Perhaps the most closely related is a recent line of work on active information acquisition, although with \emph{fixed} sensors onboard robots:
Atanasov et al. propose a non-greedy, value-iteration based offline solution with applications to gas distribution mapping and target localization~\cite{atanasov2014information}.
Schlotfeldt et al. then reformulate the problem as a deterministic planning problem and apply \algo{A*} search with the first consistent heuristic for information acquisition, with applications to active mapping~\cite{schlotfeldt2019maximum}.
Kantaros et al. then propose a probabilistically complete and asymptotically optimal sampling-based approach to this problem, along with strategies to bias exploration toward informative regions~\cite{kantaros2019asymptotically}.
Lu et al. propose a potential-function based method for integrated planning and control of robotic sensors deployed to classify multiple targets in an obstacle-populated environment~\cite{lu2014information}.

There are also lines of work that formulate information gathering as an Orienteering Problem.
Of particular interest are~\cite{vavna2015dubins,faigl2017solution,pvenivcka2019physical,pvenivcka2019data} because of a similarity in their approaches with our initialize-and-refine approach in \Split{} (detailed further in Sec.~\ref{sec:approach-two}).
However these approaches focus on computing informative \emph{tours}\tem unlike our goal-directed setting, and perform local refinement over heading angles constrained by Dubin's-car dynamics\tem unlike our approach that refines sensor angles that observe the environment.
\section{Problem Formulation and Notation}\label{sec:problem-and-notation}
%
%
\subsection{Persistent coverage framework}
We contextualize and evaluate our approaches within the persistent-coverage framework established in previous work~\cite{kusnur2019planning}.
This is a centralized framework that continuously computes goal locations to which UAVs should fly and kinodynamically feasible, globally deconflicting plans for them to do so, in a \emph{prioritized planning} setting~\cite{erdmann1987multiple}.
While it is a multi-UAV system, we plan for UAVs independently\tem plans between UAVs are not explictly coordinated in~\cite{kusnur2019planning} and is out of the scope of this paper.
Our specific contribution lies in planning robot and sensor trajectories for a single UAV navigating to a goal.
The framework in~\cite{kusnur2019planning} assumes a circular sensor footprint directly underneath the UAV.
In this paper, we extend the system to incorporate a rectangular footprint offset from the UAV\tem consequently, different sensor headings correspond to the UAV observing different areas of the environment around it.

\noindent \textbf{Map.}
The environment map $\Mmap$ consists of a priority map $\Pmap$ and a no-coverage map $\Nmap$.
The UAV must attempt to cover each cell $c_{i,j}$ at row $i$ and column $j$ of $\Pmap$.
Such a cell is associated with two values: its lifetime $l(i,j)$ and age $a(i,j)$\tem the age of a cell is the time passed since the cell was last covered by a UAV, and its lifetime is a desired bound on its age.
At any point of time t, $\Pmap$ holds the quantity $p(i,j) = l(i,j) - a(i,j)$ for each cell $c_{i,j}$.
$\Pmap$ \emph{decays} with time, meaning $p(i,j)$ for each cell $c_{i,j}$ reduces by one every second, thus making $c_{i,j}$ more \emph{urgent}.
Cells part of $\Nmap$ do not need to be covered.

\noindent \textbf{Sensor.}
In our setting, the sensor is a pan-only camera with one controllable DoF (yaw), which controls a downward-looking rectangular footprint of fixed and limited field-of-view.
The area of the footprint is discretized into cells on the map according to an underlying resolution.
We assume no noise in the footprint observed by the sensor.

\noindent \textbf{Robot.}
The UAV is a kinodynamically constrained system, accounting for the robot's $x$ and $y$ coordinates, heading angle $\theta$, velocity $v$, and timestamp $t$.
The UAV is said to be at a cell $c_{i,j}$ if the projection of its reference point onto the \xyplane{} lies
in cell $c_{i,j}$.
A cell is said to be covered by the UAV if any point on the cell is contained in the rectangular projection of the sensor footprint on the \xyplane{}.

\subsection{Problem formulation and definitions}\label{problem}
We represent this planning problem as a search over a finite, discrete search space.
Here, we define the configuration spaces of the robot (UAV) and sensor, and three state spaces that are relevant to our approaches.
Each state space is associated with a set of transitions, and they together define three separate search spaces.
\begin{figure}[t]
    \centering
    \includegraphics[width=.9\columnwidth]{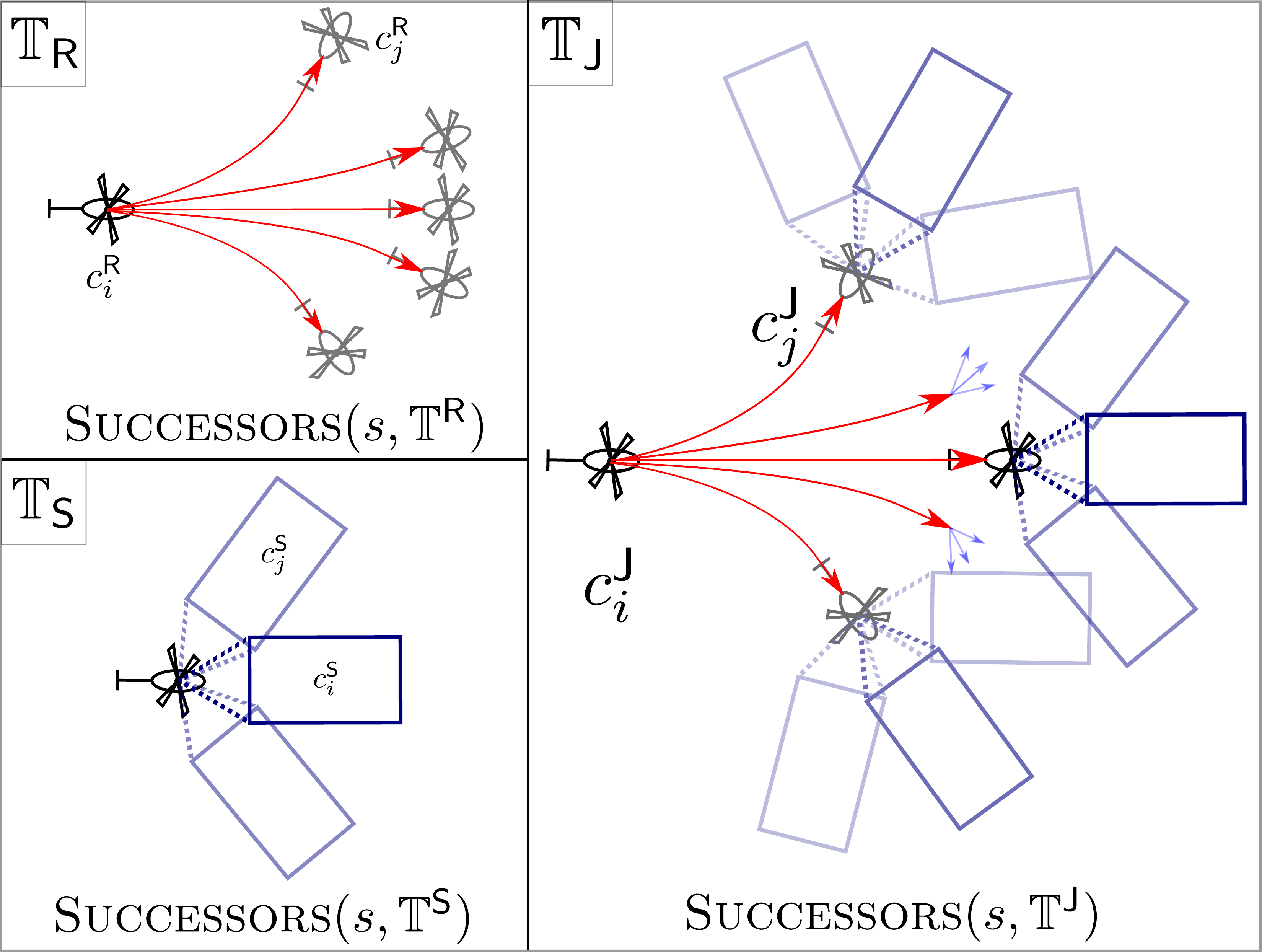}
    \caption{An example of the successor-generation functions for the three search spaces described in Sec.~\ref{sec:problem-and-notation}.}
    \label{fig:transitions}
\end{figure}
    \subsubsection{Robot state space}\label{def:robot-space}
    A feasible robot configuration is represented by $\Crobot = (x, y, \theta, v, t)$ where $x$ and $y$ are the robot's $2D$ coordinates, $\theta$ is the UAV's heading, and $t$ is the global timestamp at which this configuration is achieved
    (the timestamp $t$ is part of $\Crobot$ as we plan spatiotemporally collision-free trajectories for multiple robots in this framework).
    These five degrees of freedom together define the $5D$ robot state space $\Erobot$.
    A set of kinodynamically feasible motion primitives computed offline define a state lattice~\cite{pivtoraiko2005generating} via a set of transitions
    $$ \TranR = \{ (\Crobot_i, \Crobot_j)\ |\ \Crobot_i, \Crobot_j \in \Erobot \} $$
    This defines a search space represented by a graph $\Grobot$ with nodes $\Erobot$ and edges $\TranR$.
    A robot trajectory $\Trobot$ is a sequence of feasible robot configurations. 
    \subsubsection{Sensor state space}\label{def:sensor-space}
    A sensor configuration is defined with respect to a corresponding robot configuration $\Crobot$ as a tuple $\Csensor = (t, \psi, \Hsensor)$, where $t$ is the timestamp in $\Crobot$, $\psi$ is the sensor's heading angle in the global frame, and $\Hsensor$ is a list denoting the history of sensor angles assigned at all timestamps earlier than $t$.
    These state variables collectively define the sensor state space $\Esensor$, with dimensionality $(1 + |\Hsensor|)$\footnote[3]{Note that $t$ is a known variable and no search is performed over it, and thus it does not contribute to the dimensionality of \Csensor.}.
    The set of feasible sensor motions define a set of transitions
    \[
        \TranS = \{ (\Csensor_i, \Csensor_j)\ |\ \Csensor_i, \Csensor_j \in \Esensor \}
    \]
    This defines a search space represented by a graph $\Gsensor$ with nodes $\Esensor$ and edges $\TranS$.
    A sensor trajectory $\Tsensor$ is a sequence of sensor configurations. 
    \subsubsection{Joint state space}\label{def:joint-space}
    A feasible \emph{joint-state} configuration $\Cjoint$ is a concatenation of a feasible robot configuration and sensor configuration $\langle \Crobot, \Csensor \rangle$.
    These state variables collectively define the joint state space $\Ejoint$ of dimensionality $(6 + |\Hsensor|)$.
    The set of feasible transitions in $\Ejoint$ is a combination of feasible transitions in $\Erobot$ and $\Esensor$
    \[
        \TranJ = \{ (\Cjoint_i, \Cjoint_j)\ |\ \Cjoint_i, \Cjoint_j \in \Ejoint \}
    \]
    This defines a search space represented by a graph $\Gjoint$ with nodes $\Ejoint$ and edges $\TranJ$.
    Note that since the state-lattice discretization in $\Erobot$ can be different from that in $\Esensor$, the transition set $\TranJ$ consists of actions that change robot and sensor states at their respective state discretizations.

	Given these search spaces, we define the routine \algo{Successors}$(s, \TranR)$ to be the successor-generation routine for a state $s$ that returns successor states in $\Erobot$.
	Similarly, we have the routines \algo{Successors}$(s, \TranS)$ and \algo{Successors}$(s, \TranJ)$.
	Fig.~\ref{fig:transitions} illustrates these three types of successors, although with much smaller branching factors for \TranR and \TranJ.
	Specifically, in \TranJ for example, we generate $3$ sensor-space successors for points at every 1s of a $4$-second long motion primitive.
	We have $12$ motion primitives per robot state on average, making the branching factor in the joint space $12 \times 4 \times 3 = 144$ on average.
	We also denote running algorithm \algo{X} searching for a path from $\sStart$ to $\sGoal$ in search space $\mathbb{G}$ by \algo{X}(\sStart,~\sGoal~$|$~$\mathbb{G}$).
    For example, \algo{A*(\sStart, \sGoal $|$ \Gjoint)} denotes running \algo{A*} search in the search space determined by \Gjoint (meaning state transitions are determined by \TranJ).

\subsection{Cost Function}\label{sec:cost-function}
We now define the cost function associated with a transition from state $s$ to $s'$.
We use two costs\tem one associated with sensor coverage at $s'$ where $s' \in \Esensor$ or $\Ejoint$, and the other associated with the UAV's motion primitive from $s$ to $s'$ where $s, s' \in \Erobot$ or $\Ejoint$.

\textbf{Motion primitive cost.}
Each motion primitive is a sequence of states forward-simulated from the corresponding robot state at $s,\ s_{robot} = (x, y, \theta, v, t)$, following double-integrator dynamics.
The cost of the primitive is equal to the time taken for the UAV to execute it.
More details about the motion primitives can be found in~\cite{kusnur2019planning}.

\textbf{Sensor coverage cost.}
For the corresponding sensor state at $s'$, $s'_{sensor} = (t, \psi, H^\psi)$, the variables $x, y, \theta, \psi$ together define a 2D specific footprint of cells $\mathcal{F}$.
Let a given footprint $\mathcal{F}$ cover $|\mathcal{F}|$ discrete cells in the map $\mathcal{M}$.
Let the number of these cells lying in a coverage zone be given by $N_{\mathcal{C}}$ and those lying in a no-coverage zone be $N_{\mathcal{NC}}$:
\[
    N_\mathcal{C} = \sum\limits_{i \in \mathcal{F}} \mathds{1}\left[i \in \Pmap\right] \ \text{and}\  N_\mathcal{NC} = \sum\limits_{i \in \mathcal{F}} \mathds{1}\left[i \in \Nmap\right]
\]

\subsubsection{No sensor history}
If we ignore the sensor history, the cost of a footprint is given by the sum of priorities of all coverage cells in $\mathcal{F}$, and an additive penalty $\lambda$ scaled by the fraction of no-coverage cells in $\mathcal{F}$:
\begin{equation}\label{eqn:cost-without-history}
\mathsf{cost}_{0}(\mathcal{F}) =
	\overbrace {
            \sum_{ i \in \mathcal{F} \land \mathcal{C} } p_i
	}^\text{criticality measure}
		+\
	\overbrace {
        \mathbf{\lambda}
        \times
            \frac{ N_{\mathcal{NC}} }{ | \mathcal{F} | }
    }^{\substack{\text{penalize} \\ \text{no-coverage cells}}}
\end{equation}

\subsubsection{With sensor history}
\begin{figure}[t]
    \centering
    \includegraphics[width=\columnwidth]{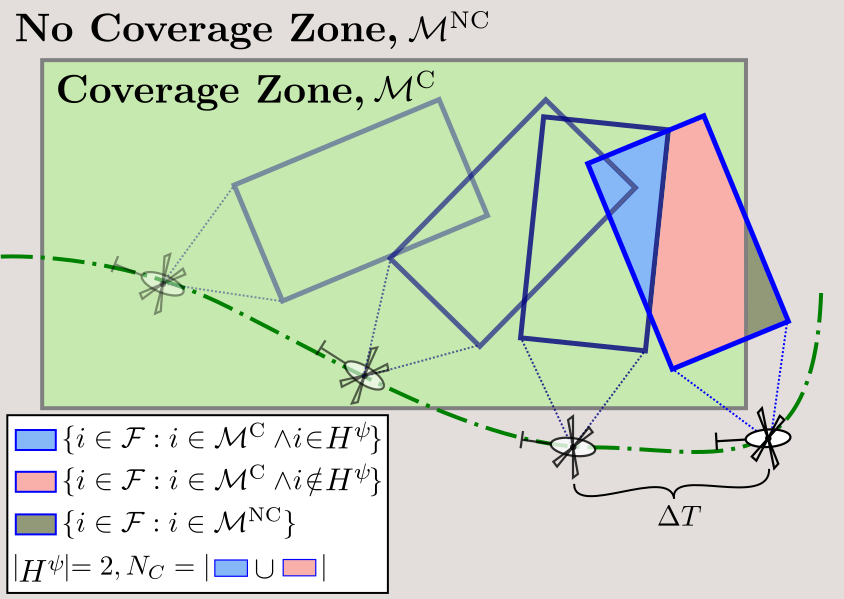}
    \caption{Pictorial explanation of our cost function $\mathsf{cost}_{H}(\mathcal{F})$ from Eq.~\ref{eqn:cost-with-history}. We consider history size, $H = 2$ in this example. For the last UAV state on the green trajectory $\Trobot$, the sensor footprint is shaded in three colours. The blue area is the overlap with previous footprints in $\Trobot$, while the red and dark green areas do not overlap. For Eq.~\ref{eqn:cost-with-history}, the blue area is \textit{in history}, the red area is \textit{not in history}, and the dark green area is \textit{penalize no-coverage cells}. Note that for footprints too far in the past, even if there was an overlap, it has no effect.}
    \label{fig:cost-function}
\end{figure}

We define the following sensor coverage cost for this footprint $\mathcal{F}$ (where `$\mathds{1}$' represents the indicator function):

{\footnotesize
\begin{equation}\label{eqn:cost-with-history}
	\mathsf{cost}_{H}(\mathcal{F}) =
		\overbrace {
            \sum_{ i \in \mathcal{F} \land \mathcal{C} }
				\underbrace { \mathds{1}[ i \notin \mathcal{H}^\psi ] \times p_i }_\text{not in history} +
				\underbrace { \mathds{1}[ i \in \mathcal{H}^\psi ] \times l_i }_\text{in history}
		}^\text{criticality measure}
				+\
            \overbrace {
                \mathbf{\lambda} \times  \frac{ N_{\mathcal{NC}} }{ | \mathcal{F} | } }^{\substack{\text{penalize} \\ \text{no-coverage cells}}}
\end{equation}
}
This cost function is illustrated in Fig.~\ref{fig:cost-function}.
Note that Eq.~\ref{eqn:cost-with-history} reduces to Eq.~\ref{eqn:cost-without-history} when no history is considered.
%
\section{\SplashVerb{}}\label{sec:approach-one}
In this section, we describe out first approach, \Splash{}.
\Splash{} first quickly computes a suboptimal robot trajectory using Multi-Heuristic \algo{A*} (\algo{MHA*}) search in \Grobot.
This search is performed with the motion primitive cost function (Sec.~\ref{sec:cost-function}).
Then, it computes a sensor trajectory using (uninformed-)\algo{A*} search in \Gsensor for a given history $\mathsf{H}$.
This search is performed with the sensor coverage cost function $\mathsf{cost}_{H} (\mathcal{F})$ (Sec.~\ref{sec:cost-function}).

\algo{MHA*} is a variant of A* that can use multiple arbitrarily inadmissible heuristics. We omit details for brevity and refer the reader to the paper for details~\cite{aine2016multi}.
We use (1) a Euclidean distance heuristic, (2) a Dubin's path length heuristic, and (3) a Dijkstra's shortest path length heuristic.

Note that here and henceforth in this paper, when we mention \algo{A*}, we are talking about \emph{uninformed} \algo{A*} (without a heuristic).
We set aside formulating with a consistent heuristic for sensor coverage in our setting for future work.
However, both \Splash{} and \Split{} both work unchanged with the addition of a heuristic.
The pseudocode for \Splash{} can be found in Alg.~\ref{alg:decoupled}.
\begin{figure}[t]
    \centering
    \includegraphics[width=\columnwidth]{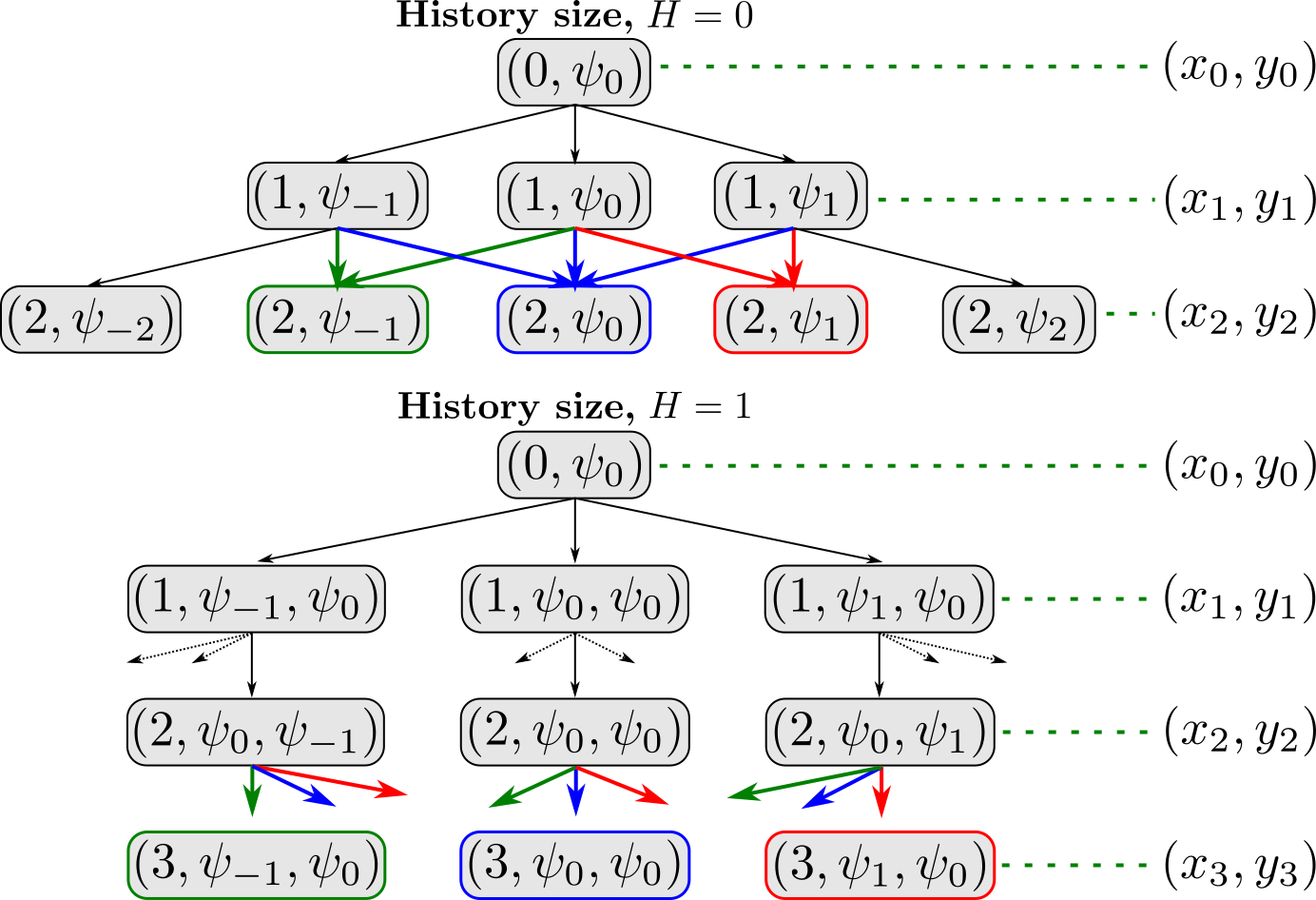}
    \caption{Graph representation for sensor planning for history size $H = 0$ (\emph{above}) and $H = 1$ (\emph{below}). Each level $l$ in the graph corresponds to a state in the UAV trajectory $\Trobot$. The search space size increases with increasing history sizes. Thus, duplicates (highlighted by coloured arrows and nodes) appear less frequently with increasing history sizes making the search for an optimal $\Tsensor$ more expensive.}
    \label{fig:history}
\end{figure}

\begin{algorithm}[h]
\footnotesize
\caption{\SplashVerb{}}\label{alg:decoupled}
  \hspace*{\algorithmicindent} \textbf{Input:} \sStart, \sGoal, $\mathsf{H}$ \\
  \hspace*{\algorithmicindent} \textbf{Output:} $\pi_f$ (final trajectory in joint space)
    \begin{algorithmic}[1]
    \Procedure{Main()}{}
        \State $\pi_\mathit{robot} \gets$ \algo{MHA*(\sStart, \sGoal $|$ \Grobot)}
        \Statex \Comment{using motion primitive cost as in Sec.~\ref{sec:cost-function}}
        \State $\pi_\mathit{sensor} \gets$ \algo{A*}(\sStart, \sGoal $|$ \Gsensor) with $\mathsf{H}$ states in sensor history
        \Statex \Comment{using sensor coverage cost, $\mathsf{cost}_{H}(\mathcal{F})$ as in Sec.~\ref{sec:cost-function}}
        \State $\pi_\mathit{joint} \gets$ concatenate $\pi_\mathit{robot}$ and $\pi_\mathit{sensor}$
        \Statex \Comment{creates a joint-space plan as in Sec.~\ref{def:joint-space}}
        \State \Return $\pi_\mathit{joint}$
    \EndProcedure
    \end{algorithmic}
\end{algorithm}

The most important aspect of \Splash{} is accounting for sensor history in Line 3.
Fig.~\ref{fig:history} illustrates the effect of history values $H = 0$ and $H = 1$ on the search graph $\Gsensor$ for a given initial sensor heading $\psi_0$.
Each level in Fig.~\ref{fig:history} corresponds to a waypoint along the robot trajectory $\Trobot$. The figure denotes state $\Csensor = (t, \psi, \Hsensor)$ as a tuple where $\Hsensor$ is the last $H$ elements in the tuple.
For any state $\Csensor$, the sensor angle can either be changed by one step (increment or decrement), or it may remain the same.

The effect of $H$ values is illustrated by the coloured arrows and vertices in the graph\tem two arrows of the same color end up at a unique state in the graph.
The key idea is as follows: For $H = 0$, ending up at $\psi_0$ on level $2$ is considered the same state, whether you come from $\psi_0$ or $\psi_{-1}$ or $\psi_{1}$\tem we only care about the \emph{current} sensor angle.
But for $H = 1$, ending up at $\psi_0$ on level $2$ is considered a different state in all these three cases because we maintain $1$ historical sensor angle.
This can be incorporated by defining the state as $\Csensor = (t, \psi, \Hsensor)$.
Observe that states are replicated in this way a lot more frequently for $H = 0$ than for $H = 1$, meaning the graph for $H = 0$ has much (in fact, exponentially) lesser states than that for $H = 1$.
\section{\SplitVerb{}}\label{sec:approach-two}
\begin{algorithm}[t]
\footnotesize
\caption{\SplitVerb{}}\label{alg:hybrid}
  \hspace*{\algorithmicindent} \textbf{Input:} \sStart, \sGoal, $\mathsf{T}_{\mathsf{overall}}$ (time limit) \\
  \hspace*{\algorithmicindent} \textbf{Output:} $\pi_f$ (final trajectory in joint space)
    \begin{algorithmic}[1]
    \Procedure{Main()}{}
        \State $\pi_i \gets$ \Splash{}$(\sStart, \sGoal, \mathsf{H}=0)$\Comment{\textbf{Initialization step}}
        \State $\mathsf{t}_\text{\Splash{}} \gets$ time taken for \Splash{} to terminate
        \State $\pi_f \gets$ \algo{LocalIterativeTunneling}($\pi, \mathsf{T}_\mathsf{overall} - \mathsf{t}_\text{\Splash{}}$)
        \Statex \Comment{\textbf{Refinement step}}
        \State \Return $\pi_f$
    \EndProcedure
    \item[]
    \Procedure{\algo{LocalIterativeTunneling}($\pi_i$, $\mathsf{t}$)}{}
    \State{\color{Blue} iteration $\gets$ 1}
    \State{\color{Blue} Create and store all states $s_i \in \pi_i$ in memory with $level(s_i) = 0$}
    \While{{\color{Blue} time $\mathsf{t}$ remains}}\Comment{{Iterative Tunneling loop}}
        \State $\sStart \gets$ first state in $\pi$
        \State $\sGoal \gets$ last state in $\pi$
        \State $g(\sGoal) = \infty$; $g(\sStart) = 0$
        \State $bp(\sStart) = bp(\sGoal) = $ \texttt{NULL}
        \State Insert $\sStart$ into \texttt{OPEN} with \algo{key}$(\sStart)$
        \While{$\text{\texttt{OPEN}}$ not empty}\Comment{{Modified \algo{A*} loop}}
            \State $s \gets$ \texttt{OPEN}.\algo{min()}\Comment{where \texttt{OPEN} is a min-heap}
            \If{$s$ is goal}
                \State Backtrack from $s$ to obtain solution $\pi_f$
                \State \textbf{break}
            \EndIf
            \For{successor $s'$ in \algo{Successors}$(s, \TranJ)$}
            \Statex \Comment{successors are computed via transitions in $\Ejoint$}
                \If{$s'$ not closed}
                    \If{$g(s') > g(s) + c(s, s')$}
                        \State $g(s') = g(s) + c(s, s')$
                        \Statex \Comment{where $c(s,s')$ is the sensor coverage cost}
                        \State $bp(s') = s $
                        \State{\color{Blue} $level(s') = level(s) + 1$}
                    \EndIf
                    \If{{\color{Blue} $level(s') \leq $ iteration}}
                        \State{\color{Blue} Insert $s'$ into \texttt{OPEN} with \algo{key}$(s')$}
                    \EndIf
                \EndIf
            \EndFor
        \EndWhile
        \State{\color{Blue} iteration $\gets$ iteration + 1}
        \State{\color{Blue} $\mathsf{t} \gets \mathsf{t}-\text{time elapsed in current iteration}$}
    \EndWhile
    \State \Return $\pi_f$
    \EndProcedure
    \item[]
    \Procedure{\algo{key}($s$)}{}
        \State \Return $g(s) + h(s)$\Comment{where $h(.)$ is a consistent heuristic}
    \EndProcedure
    \end{algorithmic}
\end{algorithm}

\Splash{} takes into account sensor history and incentivizes the search to compute plans where overlaps are minimized.
However, it operates with a fixed, suboptimal robot trajectory that optimizes only motion primitive cost.
Recall that it is empirically observed that approximate search algorithms tend to overlook better solutions that are actually ``close'' to the computed solution in the space of solution paths~\cite{furcy2006itsa}.
The final solution that \Splash{} gives us\tem say $\pi$\tem is most likely suboptimal with respect to the coverage cost in the space of joint-space solutions.
\SplitVerb{} locally refines $\pi$ by performing searches in small search spaces around $\pi$ in the joint space, increasing in size with each iteration.
We call these search spaces \emph{tunnels}, and this is an application of the ITSA* algorithm~\cite{furcy2006itsa}.

We provide pseudocode for \Split{} in Alg.~\ref{alg:hybrid}.
Lines in \textcolor{blue}{blue} indicate the differences from standard \algo{A*} search.
Line 2 obtains the initial solution from \Splash{}.
Then, \algo{LocalIterativeTunneling} refines this solution locally by performing \algo{A*} searches in iterative tunnels.
The ``level'' of a state $s$ corresponds to the distance from the initial path $\pi_i$ to $s$ computed as the smallest number of edges on a path from any state on $\pi_i$ to $s$~\cite{furcy2006itsa}.
In the beginning, every state on the initial plan $\pi_i$ is stored in memory with level $0$.

The refinement process is essentially \algo{A*} being performed repeatedly, with the addition of lines $25$\textendash $27$.
The level of any newly generated state is set to one more than the level of its parent.
Only a state whose level is lesser than the current iteration number is inserted into the OPEN list.
This is what creates tunnels increasing in size per iteration.
\algo{LocalIterativeTunneling}\tem and consequently, \Split{}\tem terminates when the time available for local refinement runs out.
\section{Experimental Results}\label{sec:results}
We evaluate our approaches by running them over $200$ randomly generated start-goal pairs over several maps.
We pick maps as seen in the persistent coverage framework described in~\cite{kusnur2019planning} (Sec.~\ref{sec:problem-and-notation}) ($10$ start-goal pairs per map for $20$ maps).
The maps are generated by letting the map in the framework decay for several minutes while one UAV with a fixed sensor covers it persistently.
We pick versions of the map at different points in time, which gives us maps with complex coverage zones.

\textbf{Evaluation.}
We use the following set of facts to compare any two trajectories:
Let a given trajectory in the joint space cover $N$ cells that lie in coverage zones.
Let the quantity $\sum_{i} p_i $ denote the sum of priorities of all such cells.
The quantity $\bar{P} = \frac{\sum_{i} p_i}{N}$ denotes the average of the priority values of all such cells.
We value two things: covering a large number of cells, and covering important cells (those with a low priority value).
Thus, a large value of $N$ and low values of $\sum_{i} p_i$ and $\bar{P}$ are desirable for a given plan.
For any two plans having the same value of $\bar{P}$, we prefer the one with the larger number of covered cells.
If a plan has a low value of $N$ as well as a large value of $\bar{P}$, it is undesirable.

\begin{figure}[t]
    \centering
    \includegraphics[width=\columnwidth]{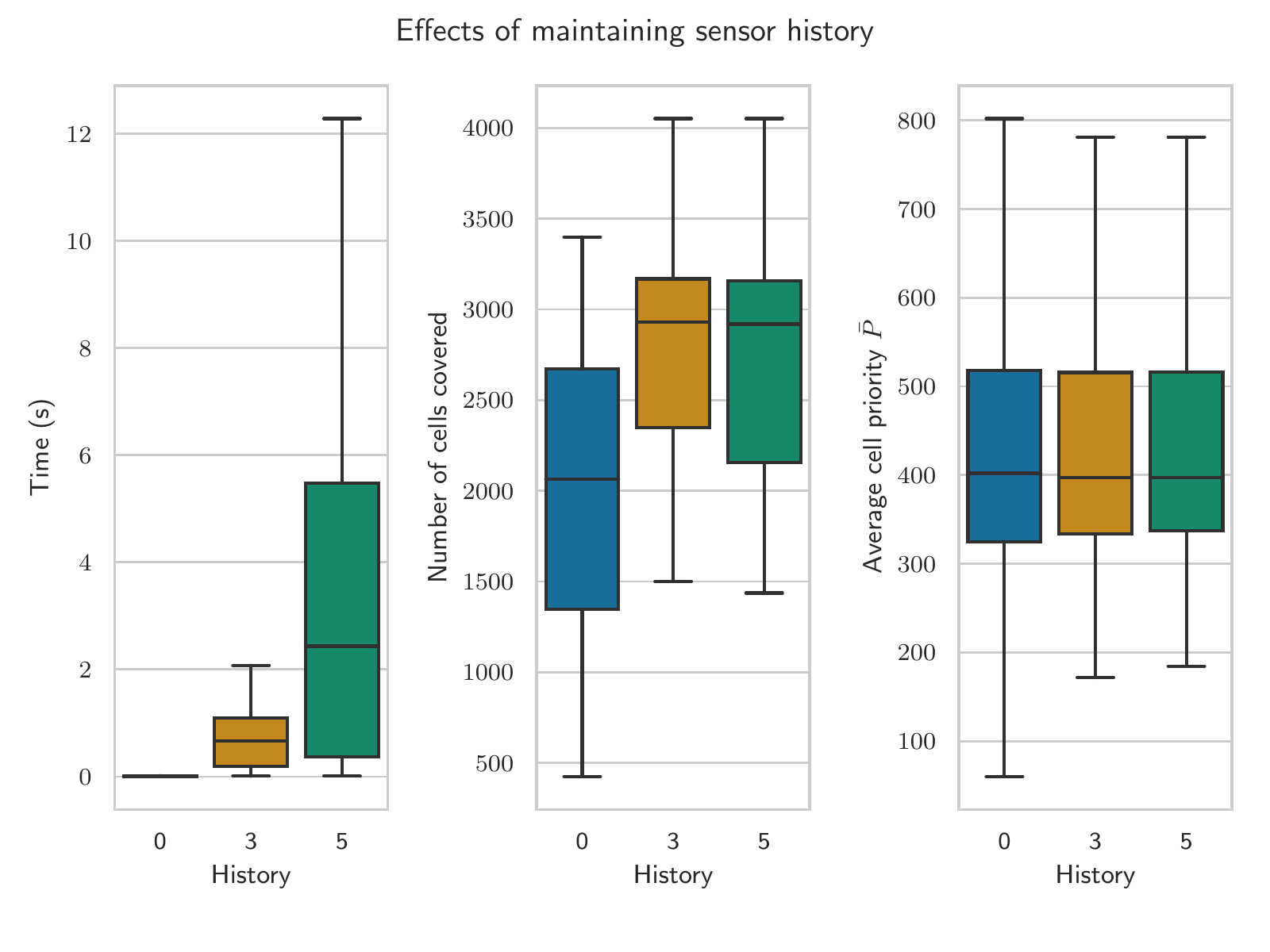}
    \caption{Results of running \Splash{} for sensor histories of size $0, 3, 5$.}
    \label{fig:splash-results}
\end{figure}

Since \Splash{} penalizes footprint overlaps, we see an increase in the number of cells covered as a larger sensor history is maintained (see Fig.~\ref{fig:splash-results}).
We also see that maintaining a sensor history of size $5$ gives us no more value than size $3$ in practical settings.
Also notice that $\bar{P}$ stays fairly unchanged over several trajectories.
This can be attributed to the maps have many different priority values for cells and no single, contiguous coverage zone\tem maintaining sensor history would lead to covering more cells, but the average priority over these cells would be approximately the same.

\begin{figure}[t]
    \centering
    \includegraphics[width=\columnwidth]{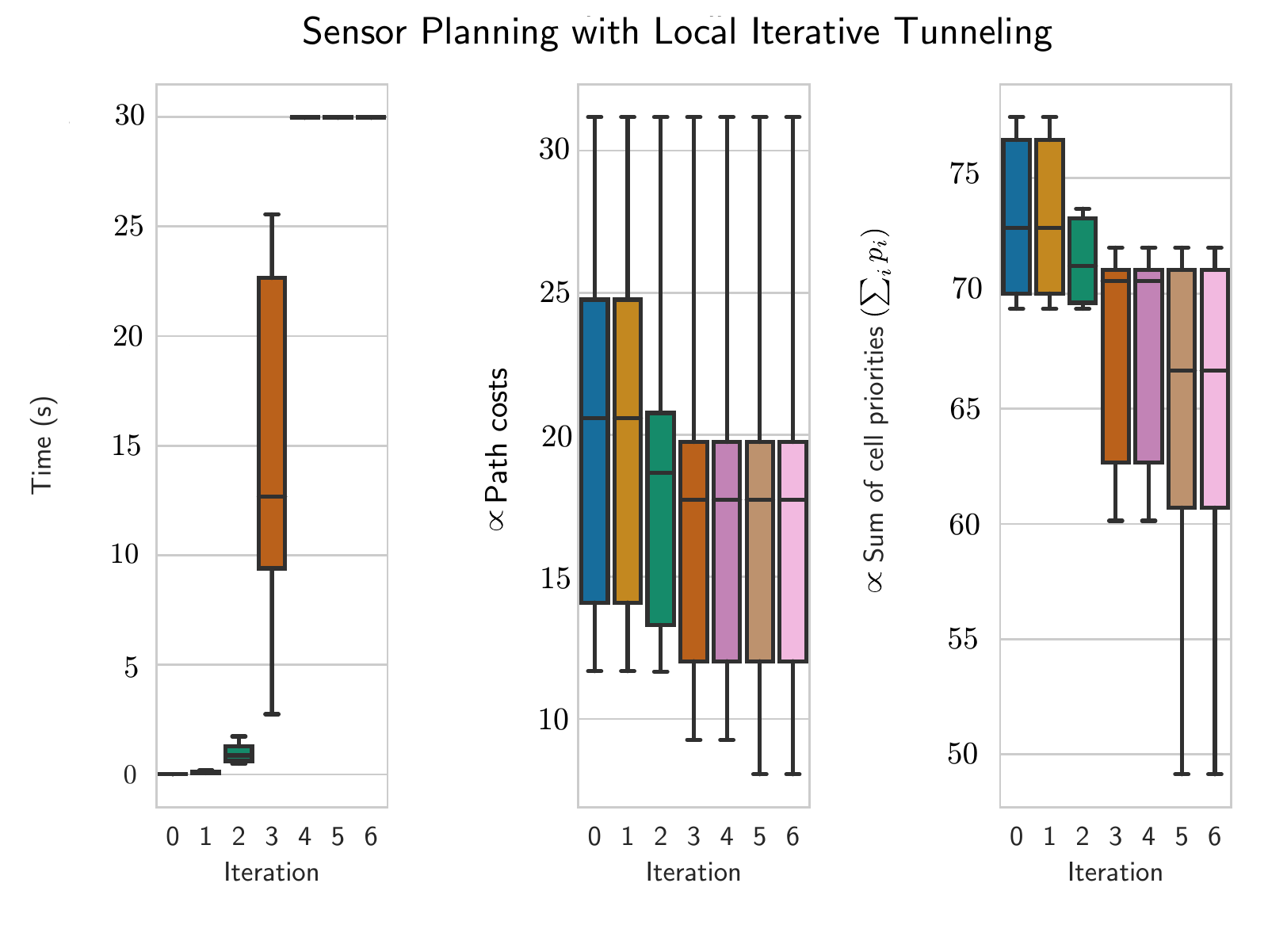}
    \caption{Results of running \Split{} timed out at $30$s.}
    \label{fig:split-results}
\end{figure}

Since \Split{} refines the trajectory locally to optimize coverage cost, we see an increase in information gain, or a decrease in $\sum_{i} p_i$, with each iteration (see Fig.~\ref{fig:split-results}).
We also see decreasing path costs (g-value of the goal) with each iteration.
If we consider a timeout of $\sim 5$s for real-time planning purposes, we see that $2$ or $3$ iterations are feasible.
Note that its performance largely depends on the immediate area around the initial plan which will be explored in the iterative tunnels.
If this immediate area has only a few more important cells to cover, the refined plan will largely stay the same.

A natural baseline is to search directly in the joint space of robot and sensor variables.
This requires a cost function that is a linear combination of the motion primitive and sensor coverage cost.
Running \algo{MHA*} on these start-goal pairs with such a cost function yielded an average planning time of $8.44\ \pm\ 6.40$s\tem significantly larger than running \Splash{} with a sensor history of size $3$ or \Split{} for $2$ or $3$ iterations.
(We set a timeout of $20$s while obtaining this value, so this is a conservative estimate and true value is in fact larger.)

Note that iteration $4$ onward \Split{} takes a lot of time to converge\tem this is not useful for real-time planning problems.
However, we include them in the results to demonstrate that the solution can still improve after the third iteration (and that a local minimum is not reached within $3$ iterations).
\section{Conclusion and Future Work}\label{conclusion}
We present two search-based approaches for generating robot and sensor trajectories in goal-directed 2D coverage tasks, namely \SplashVerb{} and \SplitVerb{}.
\Splash{} solves for robot and sensor trajectories independently in decoupled state spaces while maintaining a history of sensor headings during the search.
\Split{} is a two-step approach that first quickly computes a solution in decoupled state spaces and then refines it by searching its local neighborhood in the joint space for a better solution.
We show that both these approaches are practical alternatives to running standard search-based planning in the full joint space of robot and sensor state variables.

A limitation of \algo{ITSA*}, and consequently of \Split{}, is that the \algo{A*} searches do not reuse any search efforts between subsequent iterations, and so each iteration takes longer than the last.
Reusing search efforts between iterations will lead to considerable speed-ups, leading to faster refinement of the trajectory.
Further, we do not look into maintaining sensor histories within \Split{}.
It can be useful to adaptively increase the size of the sensor history maintained with increasing iterations in \Split{}.
We set aside these two limitations as opportunities for future work.
\bibliographystyle{IEEEtran}
\bibliography{IEEEabrv,biblio}

\begin{thebibliography}{10}
\providecommand{\url}[1]{#1}
\csname url@samestyle\endcsname
\providecommand{\newblock}{\relax}
\providecommand{\bibinfo}[2]{#2}
\providecommand{\BIBentrySTDinterwordspacing}{\spaceskip=0pt\relax}
\providecommand{\BIBentryALTinterwordstretchfactor}{4}
\providecommand{\BIBentryALTinterwordspacing}{\spaceskip=\fontdimen2\font plus
\BIBentryALTinterwordstretchfactor\fontdimen3\font minus
  \fontdimen4\font\relax}
\providecommand{\BIBforeignlanguage}[2]{{%
\expandafter\ifx\csname l@#1\endcsname\relax
\typeout{** WARNING: IEEEtran.bst: No hyphenation pattern has been}%
\typeout{** loaded for the language `#1'. Using the pattern for}%
\typeout{** the default language instead.}%
\else
\language=\csname l@#1\endcsname
\fi
#2}}
\providecommand{\BIBdecl}{\relax}
\BIBdecl

\bibitem{galceran2013survey}
E.~Galceran and M.~Carreras, ``A survey on coverage path planning for
  robotics,'' \emph{Robotics and Autonomous Systems}, vol.~61, no.~12, pp.
  1258--1276, 2013.

\bibitem{nedjati2016complete}
A.~Nedjati, G.~Izbirak, B.~Vizvari, and J.~Arkat, ``Complete coverage path
  planning for a multi-{UAV} response system in post-earthquake assessment,''
  \emph{Robotics}, vol.~5, no.~4, p.~26, 2016.

\bibitem{smith2011persistent}
R.~N. Smith, M.~Schwager, S.~L. Smith, B.~H. Jones, D.~Rus, and G.~S. Sukhatme,
  ``Persistent ocean monitoring with underwater gliders: Adapting sampling
  resolution,'' \emph{{JFR}}, vol.~28, no.~5, pp. 714--741, 2011.

\bibitem{srinivasan2004airborne}
S.~Srinivasan, H.~Latchman, J.~Shea, T.~Wong, and J.~McNair, ``Airborne traffic
  surveillance systems: video surveillance of highway traffic,'' in
  \emph{Proceedings of the ACM 2nd international workshop on Video surveillance
  \& sensor networks}, 2004, pp. 131--135.

\bibitem{teixeira2018autonomous}
L.~Teixeira, I.~Alzugaray, and M.~Chli, ``Autonomous aerial inspection using
  visual-inertial robust localization and mapping,'' in \emph{{FSR}}.\hskip 1em
  plus 0.5em minus 0.4em\relax Springer, 2018, pp. 191--204.

\bibitem{pohl1970heuristic}
I.~Pohl, ``Heuristic search viewed as path finding in a graph,''
  \emph{Artificial intelligence}, vol.~1, no. 3-4, pp. 193--204, 1970.

\bibitem{furcy2006itsa}
D.~Furcy, ``{ITSA*}: Iterative tunneling search with {A*},'' in
  \emph{Proceedings of the National Conference on Artificial Intelligence
  (AAAI), Workshop on Heuristic Search, Memory-Based Heuristics and Their
  Applications, Boston, Massachusetts}, 2006.

\bibitem{kusnur2019planning}
T.~Kusnur, S.~Mukherjee, D.~M. Saxena, T.~Fukami, T.~Koyama, O.~Salzman, and
  M.~Likhachev, ``A planning framework for persistent, multi-uav coverage with
  global deconfliction,'' in \emph{2019 International Conference on Field and
  Service Robotics (FSR)}, 2019.

\bibitem{hero2011sensor}
A.~O. Hero and D.~Cochran, ``Sensor management: Past, present, and future,''
  \emph{IEEE Sensors Journal}, vol.~11, no.~12, pp. 3064--3075, 2011.

\bibitem{gupta2006stochastic}
V.~Gupta, T.~H. Chung, B.~Hassibi, and R.~M. Murray, ``On a stochastic sensor
  selection algorithm with applications in sensor scheduling and sensor
  coverage,'' \emph{Automatica}, vol.~42, no.~2, pp. 251--260, 2006.

\bibitem{huang2017visual}
A.~S. Huang, A.~Bachrach, P.~Henry, M.~Krainin, D.~Maturana, D.~Fox, and
  N.~Roy, ``Visual odometry and mapping for autonomous flight using an rgb-d
  camera,'' in \emph{Robotics Research}.\hskip 1em plus 0.5em minus 0.4em\relax
  Springer, 2017, pp. 235--252.

\bibitem{furgale2010visual}
P.~Furgale and T.~D. Barfoot, ``Visual teach and repeat for long-range rover
  autonomy,'' \emph{Journal of Field Robotics}, vol.~27, no.~5, pp. 534--560,
  2010.

\bibitem{scherer2012autonomous}
S.~Scherer, L.~Chamberlain, and S.~Singh, ``Autonomous landing at unprepared
  sites by a full-scale helicopter,'' \emph{Robotics and Autonomous Systems},
  vol.~60, no.~12, pp. 1545--1562, 2012.

\bibitem{du1999centroidal}
Q.~Du, V.~Faber, and M.~Gunzburger, ``Centroidal voronoi tessellations:
  Applications and algorithms,'' \emph{SIAM review}, vol.~41, no.~4, pp.
  637--676, 1999.

\bibitem{mori1990active}
H.~Mori, ``Active sensing in vision-based stereotyped motion,'' in \emph{EEE
  International Workshop on Intelligent Robots and Systems, Towards a New
  Frontier of Applications}.\hskip 1em plus 0.5em minus 0.4em\relax IEEE, 1990,
  pp. 167--174.

\bibitem{costante2016perception}
G.~Costante, C.~Forster, J.~Delmerico, P.~Valigi, and D.~Scaramuzza,
  ``Perception-aware path planning,'' \emph{arXiv preprint arXiv:1605.04151},
  2016.

\bibitem{meier1967optimal}
L.~Meier, J.~Peschon, and R.~Dressler, ``Optimal control of measurement
  subsystems,'' \emph{IEEE Transactions on Automatic Control}, vol.~12, no.~5,
  pp. 528--536, 1967.

\bibitem{oshman1994optimal}
Y.~Oshman, ``Optimal sensor selection strategy for discrete-time state
  estimators,'' \emph{IEEE Transactions on Aerospace and Electronic Systems},
  vol.~30, no.~2, pp. 307--314, 1994.

\bibitem{mukai1996active}
T.~Mukai and M.~Ishikawa, ``An active sensing method using estimated errors for
  multisensor fusion systems,'' \emph{IEEE Transactions on Industrial
  Electronics}, vol.~43, no.~3, pp. 380--386, 1996.

\bibitem{chung2004decentralized}
T.~H. Chung, V.~Gupta, J.~W. Burdick, and R.~M. Murray, ``On a decentralized
  active sensing strategy using mobile sensor platforms in a network,'' in
  \emph{2004 43rd IEEE Conference on Decision and Control (CDC)(IEEE Cat. No.
  04CH37601)}, vol.~2.\hskip 1em plus 0.5em minus 0.4em\relax IEEE, 2004, pp.
  1914--1919.

\bibitem{kagami2006sensor}
S.~Kagami and M.~Ishikawa, ``A sensor selection method considering
  communication delays,'' \emph{Electronics and Communications in Japan (Part
  III: Fundamental Electronic Science)}, vol.~89, no.~5, pp. 21--31, 2006.

\bibitem{bourgault2003coordinated}
F.~Bourgault, T.~Furukawa, and H.~F. Durrant-Whyte, ``Coordinated decentralized
  search for a lost target in a bayesian world,'' in \emph{Proceedings 2003
  IEEE/RSJ International Conference on Intelligent Robots and Systems (IROS
  2003)(Cat. No. 03CH37453)}, vol.~1.\hskip 1em plus 0.5em minus 0.4em\relax
  IEEE, 2003, pp. 48--53.

\bibitem{ryan2010particle}
A.~Ryan and J.~K. Hedrick, ``Particle filter based information-theoretic active
  sensing,'' \emph{Robotics and Autonomous Systems}, vol.~58, no.~5, pp.
  574--584, 2010.

\bibitem{arora2015pasp}
S.~Arora and S.~Scherer, ``Pasp: Policy based approach for sensor planning,''
  in \emph{2015 IEEE International Conference on Robotics and Automation
  (ICRA)}.\hskip 1em plus 0.5em minus 0.4em\relax IEEE, 2015, pp. 3479--3486.

\bibitem{atanasov2014information}
N.~Atanasov, J.~Le~Ny, K.~Daniilidis, and G.~J. Pappas, ``Information
  acquisition with sensing robots: Algorithms and error bounds,'' in \emph{2014
  IEEE International Conference on Robotics and Automation (ICRA)}.\hskip 1em
  plus 0.5em minus 0.4em\relax IEEE, 2014, pp. 6447--6454.

\bibitem{schlotfeldt2019maximum}
B.~Schlotfeldt, N.~Atanasov, and G.~J. Pappas, ``Maximum information bounds for
  planning active sensing trajectories,'' in \emph{2019 IEEE/RSJ International
  Conference on Intelligent Robots and Systems (IROS)}.\hskip 1em plus 0.5em
  minus 0.4em\relax IEEE, 2019, pp. 4913--4920.

\bibitem{kantaros2019asymptotically}
Y.~Kantaros, B.~Schlotfeldt, N.~Atanasov, and G.~J. Pappas, ``Asymptotically
  optimal planning for non-myopic multi-robot information gathering.'' in
  \emph{Robotics: Science and Systems}, 2019.

\bibitem{lu2014information}
W.~Lu, G.~Zhang, and S.~Ferrari, ``An information potential approach to
  integrated sensor path planning and control,'' \emph{IEEE Transactions on
  Robotics}, vol.~30, no.~4, pp. 919--934, 2014.

\bibitem{vavna2015dubins}
P.~V{\'a}{\v{n}}a and J.~Faigl, ``On the dubins traveling salesman problem with
  neighborhoods,'' in \emph{2015 IEEE/RSJ International Conference on
  Intelligent Robots and Systems (IROS)}.\hskip 1em plus 0.5em minus
  0.4em\relax IEEE, 2015, pp. 4029--4034.

\bibitem{faigl2017solution}
J.~Faigl, P.~V{\'a}{\v{n}}a, M.~Saska, T.~B{\'a}{\v{c}}a, and V.~Spurn{\`y},
  ``On solution of the dubins touring problem,'' in \emph{2017 European
  Conference on Mobile Robots (ECMR)}.\hskip 1em plus 0.5em minus 0.4em\relax
  IEEE, 2017, pp. 1--6.

\bibitem{pvenivcka2019physical}
R.~P{\v{e}}ni{\v{c}}ka, J.~Faigl, and M.~Saska, ``Physical orienteering problem
  for unmanned aerial vehicle data collection planning in environments with
  obstacles,'' \emph{IEEE Robotics and Automation Letters}, vol.~4, no.~3, pp.
  3005--3012, 2019.

\bibitem{pvenivcka2019data}
R.~P{\v{e}}ni{\v{c}}ka, J.~Faigl, M.~Saska, and P.~V{\'a}{\v{n}}a, ``Data
  collection planning with non-zero sensing distance for a budget and curvature
  constrained unmanned aerial vehicle,'' \emph{Autonomous Robots}, vol.~43,
  no.~8, pp. 1937--1956, 2019.

\bibitem{erdmann1987multiple}
M.~Erdmann and T.~Lozano-Perez, ``On multiple moving objects,''
  \emph{Algorithmica}, vol.~2, no. 1-4, p. 477, 1987.

\bibitem{pivtoraiko2005generating}
M.~Pivtoraiko and A.~Kelly, ``Generating near minimal spanning control sets for
  constrained motion planning in discrete state spaces,'' in \emph{{IROS}},
  2005, pp. 3231--3237.

\bibitem{aine2016multi}
S.~Aine, S.~Swaminathan, V.~Narayanan, V.~Hwang, and M.~Likhachev,
  ``Multi-heuristic {A*},'' \emph{The International Journal of Robotics
  Research}, vol.~35, no. 1-3, pp. 224--243, 2016.

\end{thebibliography}
\end{document}